\documentclass[sigconf]{acmart}
\usepackage{spverbatim}
\usepackage{pifont}
\usepackage{graphicx}

\AtBeginDocument{%
  }

\copyrightyear{2024}
\acmYear{2024}
\setcopyright{rightsretained}
\acmConference[UIST Adjunct '24]{The 37th Annual ACM Symposium on User Interface Software and Technology}{October 13--16, 2024}{Pittsburgh, PA, USA}
\acmBooktitle{The 37th Annual ACM Symposium on User Interface Software and Technology (UIST Adjunct '24), October 13--16, 2024, Pittsburgh, PA, USA}
\acmDOI{10.1145/3672539.3686759}
\acmISBN{979-8-4007-0718-6/24/10}
\begin{document}

\title{Towards an LLM-Based Speech Interface for Robot-Assisted Feeding}

\author{Jessie Yuan}
\affiliation{
  \institution{Carnegie Mellon University}
  \city{Pittsburgh}
  \state{PA}
  \country{USA}
}
\email{jzyuan@andrew.cmu.edu}

\author{Janavi Gupta}
\affiliation{
  \institution{Carnegie Mellon University}
  \city{Pittsburgh}
  \state{PA}
  \country{USA}
}
\email{janavig@andrew.cmu.edu}

\author{Akhil Padmanabha}
\affiliation{
  \institution{Carnegie Mellon University}
  \city{Pittsburgh}
  \state{PA}
  \country{USA}
}
\email{akhilpad@andrew.cmu.edu}

\author{Zulekha Karachiwalla}
\affiliation{
  \institution{Carnegie Mellon University}
  \city{Pittsburgh}
  \state{PA}
  \country{USA}
}
\email{zkarachi@andrew.cmu.edu }

\author{Carmel Majidi}
\affiliation{
  \institution{Carnegie Mellon University}
  \city{Pittsburgh}
  \state{PA}
  \country{USA}
}
\email{cmajidi@andrew.cmu.edu }

\author{Henny Admoni}
\affiliation{
  \institution{Carnegie Mellon University}
  \city{Pittsburgh}
  \state{PA}
  \country{USA}
}
\email{henny@cmu.edu}

\author{Zackory Erickson}
\affiliation{
  \institution{Carnegie Mellon University}
  \city{Pittsburgh}
  \state{PA}
  \country{USA}
}
\email{zackory@cmu.edu}

\renewcommand{\shortauthors}{Yuan et al.}

\begin{abstract}
Physically assistive robots present an opportunity to significantly increase the well-being and independence of individuals with motor impairments or other forms of disability who are unable to complete activities of daily living (ADLs). Speech interfaces, especially ones that utilize Large Language Models (LLMs), can enable individuals to effectively and naturally communicate high-level commands and nuanced preferences to robots. In this work, we demonstrate an LLM-based speech interface for a commercially available assistive feeding robot. Our system is based on an iteratively designed framework, from the paper ``VoicePilot: Harnessing LLMs as Speech Interfaces for Physically Assistive Robots,'' that incorporates human-centric elements for integrating LLMs as interfaces for robots. It has been evaluated through a user study with 11 older adults at an independent living facility. Videos are located on our project website\footnote{\url{https://sites.google.com/andrew.cmu.edu/voicepilot/}}
\end{abstract}


\begin{teaserfigure}
\centering
  \includegraphics[scale=0.23]
  {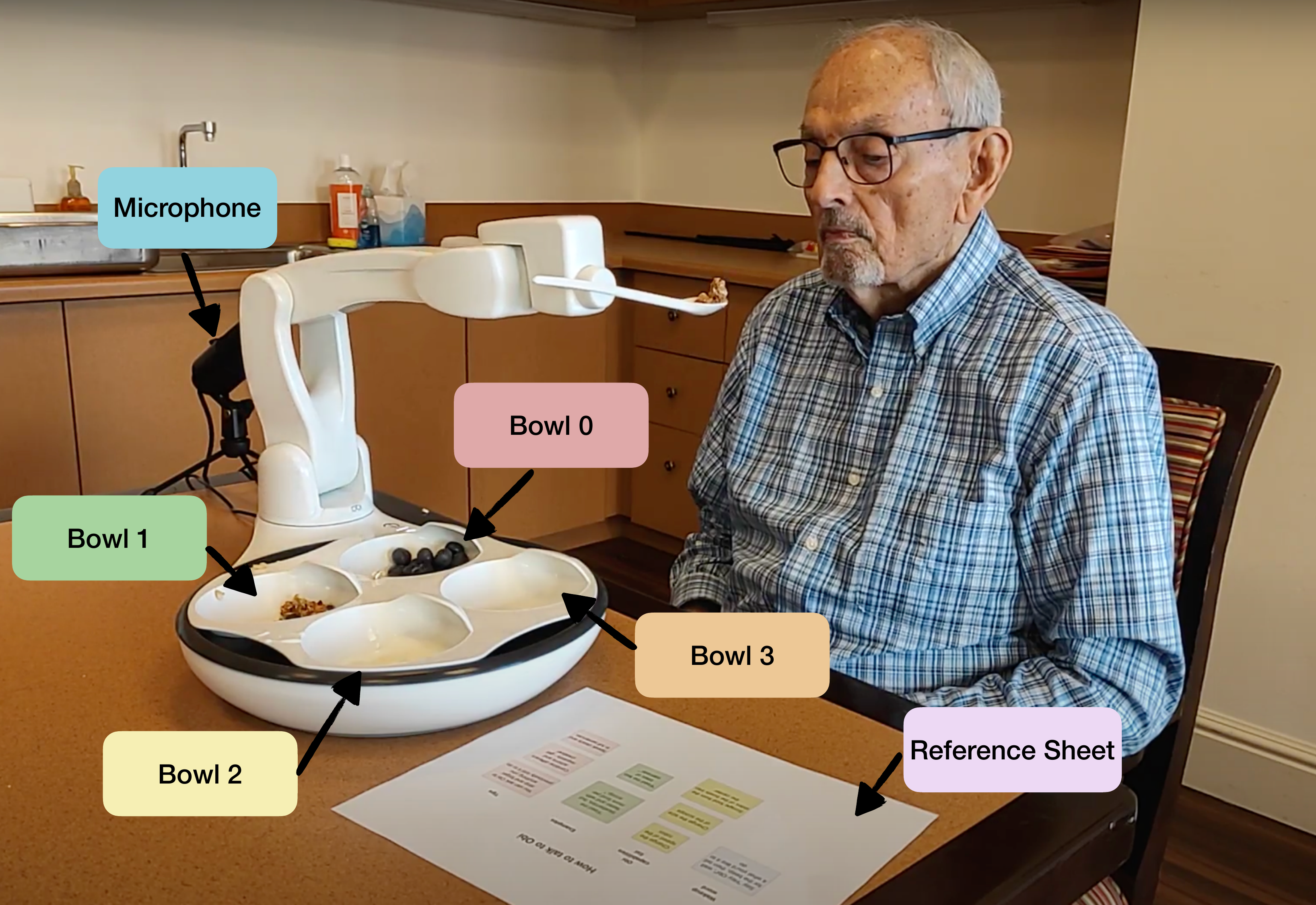}
  \vspace{-0.2cm}
  \caption{Obi robot and setup for study with 11 older adults at an independent living facility. The Obi robot's arm moves towards a participant's mouth with a spoonful of granola. A microphone is positioned to the right of the participant and a cheat sheet with example commands is placed to the left of the participant.}
  \Description{An older adult from our human study at an independent living facility is shown with the Obi robot bringing a spoonful of granola towards his mouth. A reference sheet and microphone are placed next to the Obi robot. The bowls of the robot are labelled. Bowl 0 contains blueberries. Bowl 1 contains granola. Bowl 2 contains yogurt. Bowl 3 is empty.}
  \label{fig:teaser}
\end{teaserfigure}


\maketitle

\section{INTRODUCTION}
Motor impairments affect a significant percentage of the United States with approximately 5 million individuals (1.7\%) affected by varying degrees of paralysis due to conditions including stroke and spinal cord injury~\cite{armour2016prevalence}. Physical impairments and other forms of disabilities can hinder individuals from performing activities of daily living (ADLs), such as eating and bathing. This can significantly impact an individual's independence and quality of life, requiring them to rely on a caregiver for assistance\cite{qol1, qol2, qo13, qol4}. Assistive interfaces can allow individuals with varying forms of disability to control physically assistive robots and thus enable them to perform a range of self-care and household tasks ~\cite{tasks, tasksaroundhead, king2012dusty, nanavati2023physically, yang2023high, padmanabha2023hat, padmanabha2024independence, robotsforhumanity, TapoMaya}. 

Among these interfaces, prior research shows that speech excels as an option for those with the ability to speak, as it allows individuals to naturally provide robots with both high-level commands and nuanced preferences or customizations ~\cite{pulikottil2018voice, lauretti2017comparative, house2009voicebot}. With the advent of Large Language Models (LLMs), there have been numerous advancements in using LLMs to control robots~\cite{firoozi2023foundation, kawaharazuka2024real, vemprala2023grid, singh2023progprompt, liang2023code, vemprala2023chatgpt, brohan2023rt, zhang2023large, zhao2023chat, ahn2024autort, wang2024lami, mahadevan2024generative}. However, these works concentrate on prompt engineering and code generation to enhance the accuracy of LLMs; no prior work involves human subjects interacting directly with an LLM-integrated robot for physically assistive tasks. 

In this demonstration, we introduce an LLM-based speech interface for a commercially available feeding robot, Obi ~\cite{Obi}. Our system allows novice users to easily interact with the Obi robot by speaking to it like a caregiver, enabling users to customize their dining experience to their preferences while prioritizing user safety. Our design is informed by an iteratively developed framework, as detailed in our published paper ``VoicePilot: Harnessing LLMs as Speech Interfaces for Physically Assistive Robots,'' published in the ACM Symposium on User Interface Software and Technology (UIST) 2024~\cite{padmanabha2024voicepilot}. This framework was initially developed using insights from existing literature and refined through feedback from lab members and participants at a disability awareness event. The final system was extensively evaluated in a user study involving 11 older adults at an independent living facility, approved by Carnegie Mellon University's Institutional Review Board~\cite{padmanabha2024voicepilot}. This demonstration will be a valuable reference for researchers and others interested in integrating LLMs as speech interfaces for physically assistive robots in a human-centric way.



\section{IMPLEMENTATION}

\subsection{Hardware}
Our system is implemented on the Obi feeding robot, a commercially available assistive robot consisting of a lightweight 6 degree-of-freedom arm mounted on a base containing 4 bowls. The spoon at the arm's end is detachable in case of collisions. 

\subsection{Software}
We crafted a tailored prompt for GPT-3.5 Turbo~\cite{gptturbo}, subsequently referred to as GPT, designed for the Obi robot. Users would say a wakeup phrase, “Hey Obi,” to indicate that they were about to begin a command, which was implemented using the Porcupine Python API from PicoVoice~\cite{porcupine}. The system would play a beep to indicate the the user could say their command to the robot, which would be recorded and transcribed to text using OpenAI's Whisper API~\cite{whisper}. The parsed speech-to-text commands were appended to the bottom of the prompt and provided to GPT, which generated Python code that could subsequently be deployed on the Obi robot. Prior user commands and GPT code responses were included with each subsequent command so that GPT would have an understanding of users' past preferences. 

Due to the delay required for Whisper to transcribe the command and for GPT to generate the response, the system would say ``Got it, processing,'' to indicate that it had heard the user's request. After this delay, it would announce ``Scooping now'' or ``Scraping now'' whenever it was about to execute the respective action so that users would not be surprised by the robot’s movement. Lastly, the robot would announce ``Ready for another command'' once it had finished executing the user's request. 

\subsection{Prompt Design}

\subsubsection{Environment Description}
Within our prompt, we included a physical description of the Obi robot, in addition to details on the task at hand, feeding the user, and the working environment, such as the foods in each of the bowls, as the Obi robot does not have a built-in perception system.

\subsubsection{Robot Functions and Applications}
We defined three high-level robotic control functions in the prompt so that the LLM could access these function names and use them to control the robot, along with a short description of what each function did and how it should be used. Each function only moved the robot along a predefined trajectory, as opposed to allowing GPT to directly dictate the joint angles of Obi's arm to prevent GPT from generating unsafe or unanticipated trajectories. These three functions were described as follows: \verb|obi.scoop_from_bowlno(bowlno)| moved the robotic arm to the specified bowl and scoops food from it; \verb|obi.move_to_mouth()| moved the robot to the user's mouth position; and \verb|obi.scrape_then_scoop_bowlno(bowlno)| moved the robot to the specified bowl, scraped food from the sides to the center of that bowl so that the robot could pick up additional food, then scooped from the same bowl. We designed the Robot API to include these three functions because they correspond to actions performed by the commercially available Obi robot. 

Additionally, we described how the functions related to one other and provided examples of how the functions should be combined: for instance, we specifically instructed GPT to pause between bites if users asked to be fed multiple times in succession; without this pause, GPT would immediately move from the user’s mouth to the next scoop, preventing the user from taking the bite properly. Though the default delay was set to 4 seconds, users could customize this delay if desired.

\subsubsection{Robot Variables}
To allow users to customize the way in which the robot moved, we provided the LLM with access to three robot variables that it could adjust according to the user's preferences: speed, acceleration, and scoop depth, along with their ranges and their default values. We found that grounding speed and acceleration on a 0-5 scale caused the LLM to produce adjustments that were more consistent and aligned with user expectations. The values generated by GPT were linearly scaled to the original ranges before the code was executed on the robot. To further ensure the robot variables lie within safe ranges, we introduced an LLM output processing step to clip speeds to ensure they are within the set bounds. This customization capability helps account for varying feeding contexts, such as user preferences and different types and consistencies of food. 




\subsubsection{User Control Functions}
We provided additional functions to the LLM to enhance the user's sense of control, an important consideration for assistive interfaces~\cite{padmanabha2024independence, bhattacharjee2020more, javdani2018shared}. We implemented three User Control Functions that GPT is able to call, giving users the ability to interrupt and halt the movement of the robot with a verbal command for any reason: 
\verb|obi.start()|, which begins or resumes execution of any robot code;
\verb|obi.stop()|, which permanently stops execution of any currently running robot code; and
\verb|obi.pause_indefinitely()|, which suspends execution of any currently running robot code.



\section{CONCLUSION}

In this work, we developed a speech interface using OpenAI’s GPT-3.5 Turbo for the Obi feeding robot. Future work should aim to improve the accuracy of the interface, such as by fine-tuning the model on ground-truth code output or testing out various other closed and open-source LLMs. 
Additionally, developing and testing a similar interface for another assistive robot, such as a mobile manipulator tasked with completing a variety of tasks in the home, would evaluate the interface's generalizability. Our system's balance of customizability and safety serves as a useful reference for researchers looking to design user-focused LLM-based speech interfaces for physically assistive robots.

\begin{acks}
This research was supported by the National Science Foundation Graduate Research Fellowship Program under Grant No. DGE1745016 and DGE2140739. We thank Jon Dekar and Mike Miedlar from DESĪN LLC for providing support with the Obi robot. We also thank the staff and residents of Baptist Providence Point independent living facility for their support of our human study.
\end{acks}

\bibliographystyle{ACM-Reference-Format}
\bibliography{references}

\end{document}